%% file: acl2023.tex
\pdfoutput=1

\documentclass[11pt]{article}

\usepackage{ACL2023}

\usepackage{times}
\usepackage{latexsym}
\usepackage{graphicx}
\usepackage{graphics}
\usepackage{CJK}
\usepackage{tabularx}
\usepackage{bm}
\usepackage{amssymb}
\usepackage{booktabs}
\usepackage{url}
\usepackage{multicol}
\usepackage{tabularx}
\usepackage{booktabs}
\usepackage{color}
\usepackage{arydshln}
\usepackage{multirow}
\usepackage{amsmath} 
\usepackage{bm}
\usepackage{subfigure}
\usepackage{amstext}
\usepackage{tikz}
\usepackage{pgfplots}
\usepackage{hyperref}
\usepackage{subfigure}
\usepackage{enumitem}
\usepackage{mathtools}
\usepackage{blindtext}
\usepackage{makecell}

\usepackage[T1]{fontenc}

\usepackage[utf8]{inputenc}

\usepackage{microtype}

\usepackage{inconsolata}

\usepackage{hyperref}

\usepackage{graphicx}
\usepackage{epigraph}

\usepackage{comment}

\input macros.tex

\title{\textit{Let's Negotiate!} A Survey of Negotiation Dialogue Systems}

\author{Haolan Zhan\textsuperscript{\rm \heart}, Yufei Wang\textsuperscript{\rm \heart}, Tao Feng\textsuperscript{\rm \heart}, Yuncheng Hua\textsuperscript{\rm \heart}, Suraj Sharma\textsuperscript{\rm \diamondsmall}, Zhuang Li\textsuperscript{\rm \heart}, \\ 
\textbf{Lizhen Qu}\textsuperscript{\rm \heart} \and \textbf{Gholamreza Haffari}\textsuperscript{\rm \heart} \\
\textsuperscript{\rm \heart} Department of Data Science \& AI, Monash University, Australia\\
\textsuperscript{\rm \diamondsmall}  California State University, Northridge, CA \\
\{firstname.lastname\}@monash.edu, ssharma30@gsu.edu\\ 
}
\pgfplotsset{compat=1.17} 
\begin{document}
\maketitle

\input{abstract.tex}

\input{1-introduction.tex}

\input{2-social_impact.tex}
\input{3-tech_background.tex}
\input{4-dataset.tex}
\input{6-evaluation_method.tex}

\input{5-modeling_method.tex}
\input{7-new_frontiers.tex}
\input{8-conclusion.tex}

\bibliography{ref,anthology}
\bibliographystyle{acl_natbib}



\end{document}

%% file: macros.tex
\newcommand{\heart}{\ensuremath\heartsuit}
\newcommand{\diamondsmall}{\ensuremath\diamondsuit}

%% file: abstract.tex
\begin{abstract}

Negotiation is one of the crucial abilities in human communication, and there has been a resurgent research interest in negotiation dialogue systems recently, which goal is to empower intelligent agents with such ability that can efficiently help humans resolve conflicts or reach beneficial agreements. Although there have been many explorations in negotiation dialogue systems, a systematic review of this task has to date remained notably absent. To this end, we aim to fill this gap by reviewing contemporary studies in the emerging field of negotiation dialogue systems, covering benchmarks, evaluations, and methodologies. Furthermore, we also discuss potential future directions, including multi-modal, multi-party, and cross-cultural negotiation scenarios. Our goal is to provide the community with a systematic overview of negotiation dialogue systems and to inspire future research.

\end{abstract}

%% file: 1-introduction.tex
\section{Introduction}


\epigraph{\textit{``Let us never negotiate out of fear. But let us never fear to negotiate.''}}{\textit{John F. Kennedy}}


Negotiation is one of the crucial abilities in  human communication that involves two or more individuals discussing goals and tactics to resolve conflicts, achieve mutual benefit, or find mutually acceptable solutions~\cite{fershtman1990importance,bazerman1993negotiating,lewicki2011essentials}. It is a common aspect of human interaction, occurring whenever people communicate in order to manage conflict or reach a compromise. Scientifically,  one of the long-term goals of dialogue research is to empower intelligent agents with such ability. Agent effectively negotiating with a human in natural language could have significant benefits in many scenarios, from bargaining prices in everyday trade-in~\cite{he2018decoupling} to high-stakes political or legal situations~\cite{basave2016study}.

\begin{figure}
    \centering  
    \includegraphics[width=0.49\textwidth]{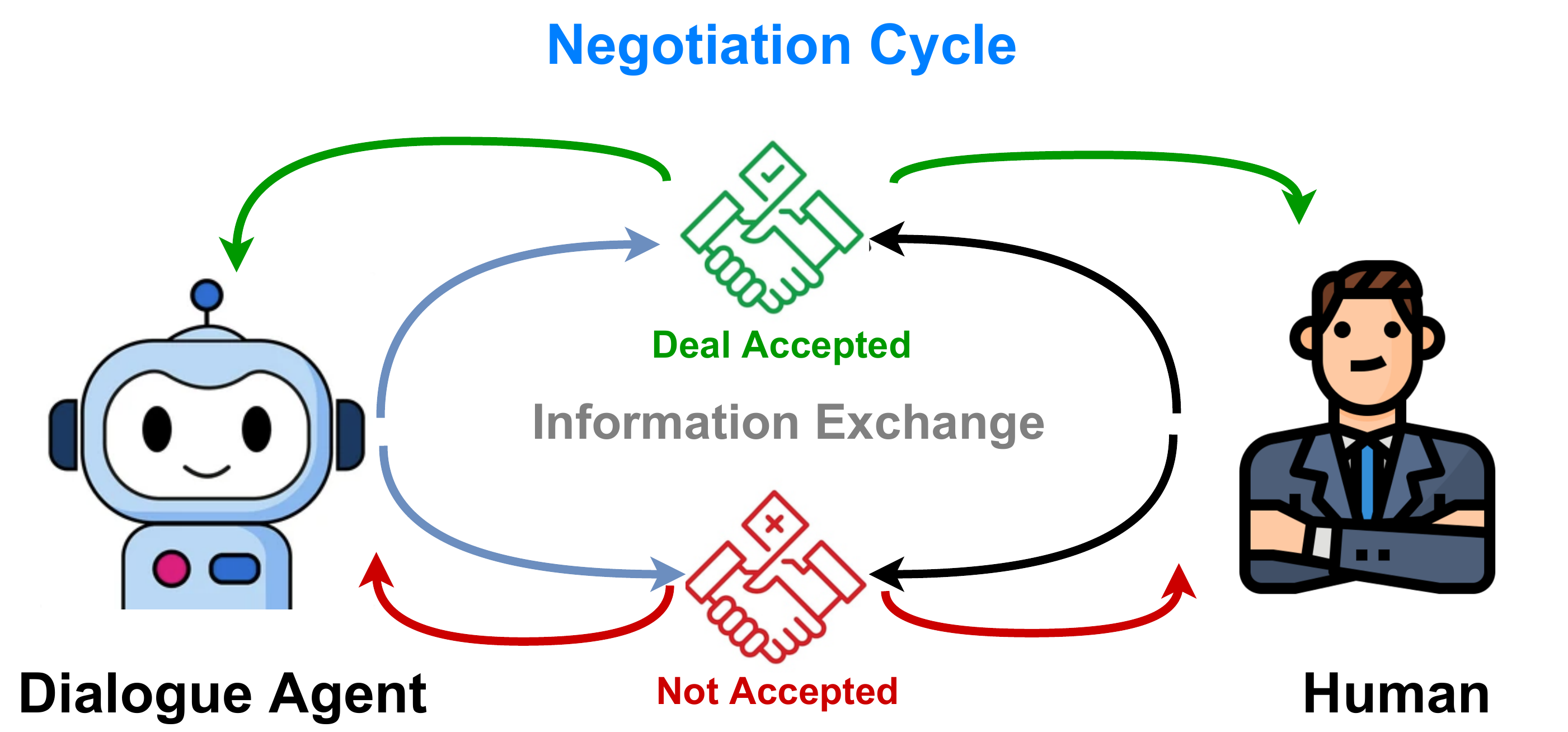}
    \caption{The negotiation process involves a multi-turn interaction between agent and human. They exchange information about their deals and end up with accepting or declining deals.}
    \label{fig:intro}
\end{figure}

Negotiation dialogue systems~\cite{lewandowska1982meaning,lambert1992modeling,chawla2021casino} is an emerging research field that aims to build intelligent conversational agents that can automatically negotiate with a human in natural languages, e.g., CICERO\footnote{\url{https://ai.facebook.com/research/cicero/}} from Meta AI. Agents negotiate with human through multi-turn interaction using logically reasoning~\cite{sycara2010agent} over  goals~\cite{zhang2020learningbebavior}, strategies~\cite{zhou2019augmenting} and psychology factors~\cite{yang2021improving}. As illustrated in Figure~\ref{fig:intro}, negotiation dialogue agents interact with the human through multi-turn cycles.
A successful negotiation process involves efficient information exchange, strategic discussion toward their goals, and a closing section.  


Despite the significant amount of research that has been conducted on the task, there is a lack of a systematic review of the topic. In this work, we aim to fill this gap by reviewing contemporary work in the emerging field of negotiation dialogue systems, covering aspects such as benchmarks, evaluation, methodology, and future directions. In recent years, various benchmarks have been proposed for negotiation dialogue systems, ranging from bargaining~\cite{lewis2017deal} and game scenarios~\cite{stac2016discourse} to job interviews~\cite{zhou-etal-2019-dynamic} and items exchanging~\cite{chawla2021casino}. 
Our survey will provide an overview of these benchmarks and discuss how they have been used to evaluate the performance of negotiation dialogue systems.

Modeling the negotiation process for conversational agents also imposes challenges. Firstly, these agents must be able to reason about and employ various strategies in different situations. In addition to strategy modeling, it is also necessary to model the personalities (e.g., mind, emotion, and behaviors) of the negotiators. Thirdly, an effective policy learning method is essential for the successful use of language. To address these challenges, we can categorize existing solutions into three areas: (1) Personality modeling helps us understand negotiator's preferences, (2) Strategy modeling enables agents to make reasonable decisions based on gathered information, and (3) Policy learning methods utilize information effectively to maximize results.

In summary, our contributions are three-fold: (1) To the best of our knowledge, we systematically categorize current negotiation dialogue benchmarks from the perspective of distributive and integrative, with each category based on different goal types of negotiation dialogue tasks. (2) We categorize typical evaluation methods and current solutions into an appropriate taxonomy. (3) We pointed out the current limitation and promising research directions in the future.








%% file: 2-social_impact.tex
\section{Backgrounds}

\subsection{Negotiation in Human}







Humans negotiate everyday in their daily routines. Negotiation is used to manage conflict and is the primary give-and-take process by which people try to reach an agreement~\cite{fisher2011getting,lewicki2011essentials}. Research on negotiation has been conducted for almost 60 years in the field of psychology, political science, and communication. It has evolved over the past decades from exploring game theory~\cite{walton1991behavioral}, behavior decisions driven by the cognitive revolution in psychology~\cite{bazerman1993negotiating}, to cultural differences in the 2000s~\cite{bazerman2000negotiation}.
Negotiation research, however, is now forced to confront the implications of human/AI collaborations given recent advancements in machine learning~\cite{sycara2010agent}.
Converging efforts from social scientists and data scientists who incorporate insights from both fields will be fruitful in maximizing expectations and outcomes during negotiation processes.

Negotiation is a process by which two or more parties attempt to resolve their opposing interests. {Strategy} of negotiation can be \textit{distributive}, such as bargaining~\cite{fershtman1990importance} and \textit{integrative}, such as maximizing unilateral interests~\cite{bazerman1993negotiating}, both of which are used in various social situations such as informal, peer to peer, organizational, and diplomatic country to country settings. The implications for enhancing outcomes are thus large and important to understand. Research from psychology demonstrates that the negotiation process can be affected by psychological factors, such as personality~\cite{sharma2013role}, relationship~\cite{olekalns2003testing}, social status~\cite{blader2012differentiating}, and cultural background~\cite{leung2011within}.  

%% file: 3-tech_background.tex






\subsection{Task Definition}

The ability of an agent to maintain good communication skills as well as strategic reasoning capabilities is what makes a negotiation dialogue system unique from typical task-oriented and open-domain systems. Negotiation dialogue aims to interact with opponents in a strategic discussion to find acceptable solutions for both parties. A negotiation process involves a mixture of strategies such as debate, persuasion, adversary, and compromise. These strategies will change along with the negotiation flow and can be influenced by opponents' personalities (e.g., thoughts, emotions, and behaviors). Therefore, \textbf{strategy} and \textbf{personality} are two main aspects to be modeled in negotiation dialogue systems. Formally, a negotiation dialogue task is defined as a tuple $(\mathcal{K}, \mathcal{S}, \mathcal{U}, \pi, {g})$. $\mathcal{K}$ refers to the pre-defined information which is prepared for the negotiation process, such as the negotiator's preferences and demands. $\mathcal{S}$ refers to a trajectory $\{s_1, s_2, ...\}$, which is used to model the strategy transition process and provided for policy learning module (e.g., reinforcement learning). A series turn of dialogue interactions $\mathcal{U}$ can be viewed as $\{u_1, u_2, ...\}$ are generated along with the negotiation process. A policy learning module $\pi_{\theta}(\mathcal{K}, \mathcal{S}, \mathcal{U})$ is utilized to learn an optimal deterministic policy which helps reach the negotiation goal $g$.



%% file: 4-dataset.tex
\section{Negotiation Datasets}
\label{datasetsection}
\begin{table*}[t]
\begin{center}
\small
\setlength{\tabcolsep}{1.6mm}{
\begin{tabular}{lccccc}
\toprule
DataSet & Negotiation Type & Scenario & \# Dialogue  & \# Avg. Turns & \# Party \\
\midrule
InitiativeTaking (\citeyear{nouri-traum-2014-initiative}) & Integrative & Fruit Assignment & 41  & - & Multi\\
STAC (\citeyear{stac2016discourse}) & Integrative & Strategy Games & 1081 & 8.5  & Two \\
DealorNoDeal (\citeyear{lewis2017deal}) & Integrative & Item Assignment & 5808 & 6.6 & Two \\
Craigslist (\citeyear{he2018decoupling}) & Distributive & Price Bargain & 6682 & 9.2 & Two \\
NegoCoach (\citeyear{zhou-etal-2019-dynamic}) & Distributive & Price Bargain & 300 & - & Two \\
PersuasionforGood (\citeyear{wang-etal-2019-persuasion}) & Distributive & Donation & 1017 & 10.43 & Two \\
FaceAct (\citeyear{dutt2020keeping}) & Distributive & Donation & 299 & 35.8 & Two\\
AntiScam (\citeyear{li2020end}) & Distributive & Privacy Protection & 220  & 12.45 & Two\\
CaSiNo (\citeyear{chawla2021casino}) & Integrative & Item Assignment & 1030 & 11.6 & Two \\
JobInterview (\citeyear{yamaguchi-etal-2021-dialogue}) & Integrative & Job Interview & 2639 & 12.7 & Two \\
DinG (\citeyear{boritchev-amblard-2022-multi}) & Integrative & Strategy Games & 10 & 2357.5 & Multi \\
\bottomrule
\end{tabular}}
\caption{Negotiation dialogues benchmarks are sorted by their publication time. For each dataset, we introduce the negotiation type, scenario, the number of dialogues and corresponding average turns, and party attributes.}
\label{tab:dataset}
\end{center}
\end{table*}
In this section, we summarize the existing negotiation datasets and resources. 
Table~\ref{tab:dataset} shows all of the collected benchmarks, along with their negotiation types, scenarios and data scale. In this paper, we categorize these benchmarks based on their negotiation types, namely, \textit{integrative} negotiation and \textit{distributive} negotiation. The integrative negotiation is associated with win-win scenarios, and its goal is to develop mutual gain. On the contrary, the distributive negotiation is often associated with win-lose scenarios and aims to maximize personal benefits. In general, The distributive negotiation is more competitive than its integrative counterpart.

\subsection{Integrative Negotiation Datasets}
In integrative negotiations, there is normally more than one issue available to be negotiated. To achieve optimal negotiation goals, the involved players should make trade-offs for multiple issues.

\paragraph{Multi-player Strategy Games} The strategy video games provide ideal platforms for people to verbally communicate with other players to accomplish their missions and goals.~\citet{stac2016discourse} propose the STAC benchmark, which is the player dialogue in the game of Catan. In this game, players need to gather resources, including wood, wheat, sheep, and more, with each other to purchase settlements, roads and cities. As each player only has access to their own resources, they have to communicate with each other. To investigate the linguistic strategies used in this situation, STAC also includes an SDRT-styled discourse structure. \citet{boritchev-amblard-2022-multi} also collect a \emph{DinG} dataset  from French-speaking players in this game. The participants are instructed to focus on the game, rather than talk about themselves. As a result, the collected dialogues can better reflect the negotiation strategy used in the game process.

\paragraph{Negotiation for Item Assignment}
The item assignment scenarios involve a fixed set of items as well as a predefined priority for each player in the dialogue. As the players only have access to their own priority, they need to negotiate with each other to exchange the items they prefer. \citet{nouri-traum-2014-initiative} propose \emph{InitiativeTalking}, occurring between the owners of two restaurants. They discuss how to distribute the fruits (i.e., apples, bananas, and strawberries) and try to reach an agreement. \citet{lewis2017deal} propose \emph{DealorNoDeal}, a similar two-party negotiation dialogue benchmark where both participants are only shown their own sets of items with a value for each and both of them are asked to maximize their total score after negotiation. \citet{chawla2021casino} propose \emph{CaSiNo}, a dataset on campsite scenarios involving campsite neighbors negotiating for additional food, water, and firewood packages. Both parties have different priorities over different items.

\paragraph{Negotiation for Job Interview}
Another commonly encountered negotiation scenario is job offer negotiation with recruiters. \citet{yamaguchi-etal-2021-dialogue} fill this gap and propose the \emph{JobInterview} dataset. JobInterview includes recruiter-applicant interactions over salary, day off, position, company, and workplace. The participants are shown negotiators’ preferences and the corresponding issues and options and are given feedback in the middle of the negotiation.

\subsection{Distributive Negotiation Datasets}
Distributive negotiation is about the discussion over a fixed amount of value (i.e., slicing up the pie). In such negotiation, the involved people normally talk about a single issue (e.g., item price) and therefore, there are hardly trade-offs between multiple issues in such negotiation. 

\paragraph{Persuasion For Donation}
Persuasion, convincing others to take specific actions, is a necessary required skill for negotiation dialogue~\cite{sycara1990persuasive,sierra1997framework}. \citet{wang-etal-2019-persuasion} focus on persuasion and propose \emph{PersuasionforGood}, a two-party persuasion conversations about charity donations. In the data annotation process, the persuaders are provided some persuasion tips and example sentences, while the persuaders are only told that this conversation is about charity. The annotators are required to complete at least ten utterances in a dialogue and are encouraged to reach an agreement at the end of the conversations. \citet{dutt2020keeping} further extend \emph{PersuasionforGood} by adding the utterance-level annotations that change the positive and/or the negative face of the participants in a conversation. A face act can either raise or attack the positive face or negative face of either the speaker or the listener in the conversation.

\paragraph{Negotiation For Product Price}
Negotiations over product prices can be observed on a daily basis. 
\citet{he2018decoupling} propose \emph{CraigslistBargain}, a negotiation benchmark based on a realistic item price bargaining scenario. In \emph{CraigslistBargain}, two agents, a buyer and a seller, are required to negotiate the price of a given item. The listing price is available to both sides, but the buyer has a private price as the target. Then two agents chat freely to decide the final price. The conversation is completed when both agents agree with the price or one of the agents quits. \citet{zhou-etal-2019-dynamic} propose \emph{NegoCoach} benchmark on similar scenarios, but with an additional negotiation coach who monitors messages between the two annotators and recommends tactics in real-time to the seller to get a better deal.

\paragraph{User Privacy Protection}
Privacy protection of negotiators has become more and more vital. Participant (e.g., attackers and defenders) goals are also conflicting. \citet{li2020end} propose \emph{Anti-Scam} benchmark which focuses on online customer service. In \emph{Anti-Scam}, users try to defend themselves by identifying whether their components are attackers who try to steal sensitive personal information. \emph{Anti-Scam} provides an opportunity to study human elicitation strategies in this scenario. 

%% file: 6-evaluation_method.tex
\section{Evaluation}

We categorize the evaluation methods for negotiation dialogue systems into three types: goal-oriented metrics, game-based metrics and human evaluation.
Table~\ref{tab:metrics} lists the evaluation metrics that are introduced in our survey. 

\subsection{Goal-based Metrics}

Goal-oriented metrics mainly consider the agent's proximity to the goal from the perspective of strategy modeling, task fulfillment and sentence realization. \textit{Success Rate (SR)} is the most widely used, which measures how frequently an agent completes the task within their goals. A similar metric \textit{Prediction Accuracy (PA)} is to evaluate the agent's strategy predictions or the outcome of negotiations, such as macro or average F1 score~\cite{wang-etal-2019-persuasion,dutt2020keeping,chawla2021casino}. For those scenario-related tasks,
\citet{yamaguchi-etal-2021-dialogue} present a task where the model is required to label the human-human negotiation outcomes as either a success or a breakdown, including area under the curve (ROC-AUC), confusion matrix (CM), and average precision (AP).  \citet{DBLP:conf/naacl/KornilovaED22} propose a model-based evaluation based on Item Response Theory to analyze the effectiveness of persuasion on the audience.

In terms of language realization for negotiation dialogue,
\citet{DBLP:books/sp/15/HiraokaNSTN15} employ a predefined naturalness metric (a bigram overlap between the system responses and the ground-truth responses) as part of the reward to evaluate policies in cooperative persuasive dialogues.
Other classical metrics for evaluating the quality of response are also used, i.e., perplexity (PPL), BLEU-2, ROUGE-L, and BOW Embedding-based Extrema matching score~\cite{lewis2017deal}.

\begin{table*}[ht]
\begin{center}
\small
\begin{tabular}{cc}
\toprule
    \makecell[c]{Goal-based \\ Metrics}  & \makecell[c]{SR, PA (\citeyear{nouri-traum-2014-initiative,wang-etal-2019-persuasion,dutt2020keeping,DBLP:conf/naacl/KornilovaED22}); Average F1 score (\citeyear{chawla2021casino}); \\ Macro F1 score (\citeyear{wang-etal-2019-persuasion,dutt2020keeping}); ROC-AUC, CM, AP (\citeyear{yamaguchi-etal-2021-dialogue});  Naturalness (\citeyear{DBLP:books/sp/15/HiraokaNSTN15}); \\ PPL, BLEU-2, ROUGE-L, Extrema (\citeyear{lewis2017deal})} \\ 
\midrule
    \makecell[c]{Game-based \\ Metrics} & \makecell[c]{WinRate, AvgVPs (\citeyear{keizer-etal-2017-evaluating}); Utility, Fairness, Length (\citeyear{he2018decoupling});\\ Average Sale-to-list Ratio, Task Completion Rate (\citeyear{zhou-etal-2019-dynamic})} \\ 
\midrule
    \makecell[c]{Human \\ Evaluation} & \makecell[c]{Agent satisfaction (\citeyear{DBLP:books/sp/15/HiraokaNSTN15, lewis2017deal}); Purchase decision, Correct response rate (\citeyear{DBLP:books/sp/15/HiraokaNSTN15}) \\ Achieved agreement rate, Pareto optimality rate (\citeyear{lewis2017deal}); Likert score (\citeyear{he2018decoupling})} \\ 
\bottomrule
\end{tabular}
\caption{Various Metrics used in the existing negotiation dialogues benchmarks.}
\label{tab:metrics}
\end{center}
\end{table*}

\subsection{Game-based Metrics}
Different from goal-oriented metrics, which focus on evaluating the accuracy of strategies or actions, game-based evaluation provides a user-centric perspective through multi-turn interactions. \citet{keizer-etal-2017-evaluating} measure the bots' negotiation strategies within the online game "Settlers of Catan" by proposing the metrics WinRate (the percentage of games won by the humans when playing with the bot opponents) and AvgVPs (the average number of victory points gained by the human players). \citet{he2018decoupling} present a task that two agents bargain to get the best deal using natural language. They use task-specific scores to test the performance of the agents, including utility (a score that is higher when the final price is closer to one agent's expected price), fairness (the difference between two agents' utilities), and length (the number of the sentences exchanged between the two agents).  \citet{zhou-etal-2019-dynamic} design a task where a seller and a buyer try to achieve a mutually acceptable price through a natural language negotiation. They adopt different metrics to evaluate the dialogue agent, i.e., average sale-to-list ratio 
and the task completion rate.
Besides, \citet{cheng2019evaluating} propose an adversarial attacking evaluation approach to test the robustness of negotiation systems.

\subsection{Human Evaluation}
To evaluate the users' satisfaction with the dialogue systems, human judgment is employed as a subjective evaluation of the generated output.
\citet{DBLP:books/sp/15/HiraokaNSTN15} use a dialogue system as the salesperson to bargain with the human customers and have the users annotate subjective customer satisfaction (a five-level score), the final decision of making a purchase (a binary number indicating whether persuasion is successful), and the correct response rate in the dialogues.
\citet{lewis2017deal} employ crowd-sourcing workers to highlight that essential information when bargaining with dialogue systems, covering the percentage of dialogues where both interlocutors finally achieve an agreement, and \emph{Pareto optimality}, i.e., the percentage of the Pareto optimal solutions in all the agreed deals. 
\citet{he2018decoupling} propose human likeness as a metric in evaluating how well the dialogue system is doing in a bargain. They ask workers to manually score the dialogue agent using a \emph{Likert} metric to judge whether the agent acts like a real human or not.


%% file: 5-modeling_method.tex
\section{Methodology Overviews}

\begin{figure}
    \centering
    \includegraphics[width=0.49\textwidth]{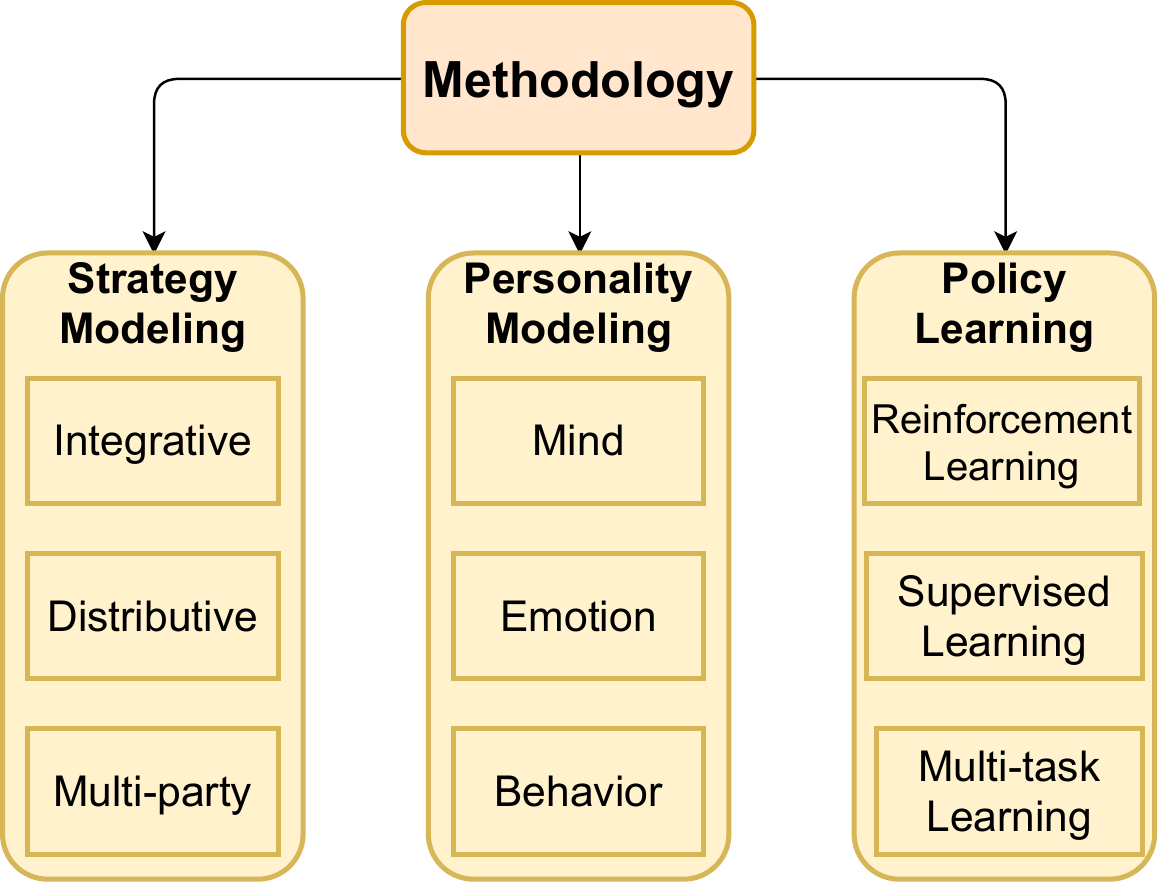}
    \caption{A summary of different methods proposed by previous research efforts for the negation dialogues.}
    \label{fig:method}
\end{figure}

As shown in Figure~\ref{fig:method}, we categorize existing methods into \emph{Strategy Modeling}, \emph{Personality Modeling}, and \emph{Policy Learning}. Strategy modeling methods help conversational agents utilize appropriate strategies in different situations. In addition, negotiation is not only about complex reasoning over strategies,  but also influenced by psychology factors. Personality is one of the most important psychology factors that would affect agents' decisions. Therefore, personality modeling helps agents perceive opponents. Besides, an effective policy
learning method is indispensable for the successful use of language.


\subsection{Strategy Modeling}


In this section, we discuss strategy modeling for negotiation dialogue systems. Negotiation strategies include a range of tactics and approaches that people use to achieve their goals in negotiation processes. It can be categorized into three aspects: \emph{integrative} (win-win) and \emph{distributive} (win-lost), and \emph{multi-party}. 

\subsubsection{Integrative Strategy}
Integrative strategy (known as \emph{win-win}) modeling aims to achieve mutual gain among participants. For instance, \citet{zhao2019rethinking} proposes to model the discourse-level strategy using a latent action reinforcement learning (LaRL) framework. LaRL can model strategy transition within a latent space. However, due to the lack of explicit strategy labels, LaRL can only analysis strategies in implicit space. To resolve the problem, \citet{chawla2021casino} define a series of explicit strategies such as \emph{Elicit-Preference}, \emph{Coordination} and \textit{Empathy}. While \textit{Elicit-Preference} is a strategy attempting to discover the preference of the opponent, \textit{Coordination}  promotes mutual benefits by explicit offer or implicit suggestion. 
In order to capture user's preference, \citet{chawla2022opponent} utilize those strategies using a hierarchical neural model. Besides, \citet{yamaguchi2021dialogue} present another collaborative strategy set to negotiate workload and salaries during the interview, which goal is to reach an agreement between employer and employee. It assists humans in becoming better negotiators during this process, e.g., communicating politely, addressing concerns, and providing side offers.


\subsubsection{Distributive Strategy}
Distributive strategy (known as \emph{win-loss}) modeling focuses on achieving one's own goals and maximizing unilateral interests more than mutual benefits. Distributive strategy can be used when you insist on your position or resist the opponent's deal~\cite{zhou-etal-2019-dynamic}. For example, \citet{dutt-etal-2021-resper} investigate four resisting categories, namely contesting, empowerment, biased processing, and avoidance \cite{fransen2015strategies}. Each individual category contains fine-grain strategic behaviors. For example, contesting refers to attacking the message source, and empowerment implies reinforcing personal preference to contradict a claim (\textit{Attitude Bolstering}) or attempting to arouse guilt in the opponent (\textit{Self Pity}). Besides, \citet{wang-etal-2019-persuasion} design a set of persuasion strategies to persuade others to donate to charity. It contains 10 different strategies containing logical appeal, emotional appeal, source-related inquiry and etc. \citet{li-etal-2020-exploring-role} explore the role structure to enhance the strategy modeling. \citet{dutt2020keeping} further enhances the role modeling with facing act, which helps utilize strategy between asymmetric roles.


\subsubsection{Multi-party Strategy}

While previously mentioned work on integrative and distributive strategy modeling mainly relates to two-party, multi-party strategy modeling is slightly different.  In multi-party situations, strategy modeling needs to consider different attitudes and complex relationships among individual participants, whole groups, and subgroups~\cite{traum2008multi}. \citet{georgila2014single} attempt to model multi-party negotiation using a multi-agent RL framework. Furthermore, \citet{shi2019deep} propose to construct a discourse dependency tree to predict relation dependency among multi-parties. Besides, \citet{li2021dadgraph} disclose relations between multi-parties using a graph neural network. However, due to the limited access to multi-party datasets, strategy modeling on multi-party scenarios is under-explored.

\subsection{Personality Modeling}

Negotiation dialogue involves complex social interactions related to multiple disciplines, such as psychology, for understanding human decision-making. Personality is an important factor in the understanding human-decision process. We summarize those work modeling personality from three perspectives: \textit{Mind}, \textit{Emotion}, and \textit{Behavior} modeling.

\subsubsection{Mind Modeling}
Mind modeling in negotiation dialogue systems encompasses several tasks, such as mind preference estimation and opponent response prediction. Mind preference estimation helps the agent infer the intention of the opponents and guess how their own utterances would affect the opponent's mental preference. \citet{nazari2015opponent} propose a heuristic frequency-based method to estimate the negotiator's preference. \citet{langlet2018detecting} consider a rule-based system incorporating linguistic features to identify user's preference. A critical challenge for mind modeling in negotiation is that it usually requires complete dialogues, so it is difficult to predict those preferences precisely for partial dialogue. To make it applicable for those partial dialogues, which is widespread in real-world applications, \citet{chawla2022opponent} formulated mind preference estimation as a ranking task and proposed a transformer-based model that can be trained directly on partial dialogue.

In terms of opponent response prediction, \citet{he2018decoupling} firstly propose to decouple the modeling of the strategy of generation containing a parser to map utterances with dialogue acts and a dialogue manager to predict the skeleton of dialogue acts. \citet{yang2021improving} further improve the negotiation system with a first-order model based on the theory of Mind~\cite{frith2005theory} , which allows the agents to compute an expected value for each mental state. They provided two variance variants of ToM-based dialogue agents: explicit and implicit, which can fit both pipeline and end-to-end systems.

\subsubsection{Emotion Modeling}

Emotion modeling refers to recognizing the emotional change between negotiators. Therefore, explicit modeling of emotions throughout a conversation is crucial to capture reflections from opponents. To study emotional feelings and expressions in negotiation dialogue, \citet{chawla2021towards} explore the prediction of two important subjective goals, including outcome satisfaction and partner perception. \citet{DBLP:conf/acl/LiuZDSLYJH20} provide explicit modeling on emotion transition engaged with pre-trained models (e.g., DialoGPT), to support help-seeker. Further, \citet{dutt2020keeping} propose a facing act modeling under the persuasive discussion scenarios. \citet{mishra2022pepds} utilize a reinforcement learning framework to engage emotion in the persuasive message. 

\subsubsection{Behavior Modeling}

Behavior modeling refers to detecting and predicting opponents' behaviors during the negotiation process. For example, fine-grained dialogue act labels are provided in the Craigslist dataset~\cite{he2018decoupling}, to help track the behaviors of buyers and sellers.  \citet{zhang2020learningbebavior} propose an opposite behavior modeling framework to estimate opposite action using DQN-based policy learning. \citet{chawla2021exploring} explore early prediction between negotiators for the outcomes. \citet{tran2022ask} leverage dialogue acts to identify optimal strategies to persuade humans for donation.

\subsection{Policy Learning}

Policy learning plays an important role in negotiation dialogue systems, by which the agent learns to choose a strategy and generate a response for the next step. Methods of policy learning can be roughly categorized into three types: \emph{reinforcement learning}, \emph{supervised learning}, and \emph{multitask learning}.


\subsubsection{Reinforcement Learning}

Reinforcement learning (RL) is one of the most common frameworks chosen for policy learning. \citet{english2005learning} are the first to use RL techniques for negotiation dialogue systems. They employed a single-agent pattern to learn the policy of two opponents individually. But single-agent RL techniques are not well suited for concurrent learning where each agent is trained against a continuously
changing environment. Therefore, \citet{georgila2014single} further advances the framework with concurrent progress using multi-agent RL techniques, which simultaneously model two parties and provide a way to deal with multi-issues scenarios. Besides, \citet{keizer-etal-2017-evaluating} propose to learn the action of the target agents with a Q-learning reward function. They further propose a method based on hand-crafted rules and a method using Random Forest trained on a large human negotiation corpus from~\cite{afantenos2012modelling}.

Most recent works try to equip RL with deep learning techniques. For instance, \citet{zhang2020learningbebavior} propose OPPA, which lets the target agent behave given the system actions. The system actions are predicted and conditioned on the target agent's actions. The reward of the actions for the target agent is obtained by predicting a structured output given the whole dialogue.
Besides, \citet{shi2021refine} use a modular framework containing a language model to generate responses, a response detector would automatically annotate the response with a negotiation strategy, and an RL-based reward function to assign a score to the strategy. Instead of the modular framework which separates policy learning and response generation, \citet{gao2021deepRL} propose an integrated framework with deep Q-learning, which includes multiple channel negotiation skills. It allows agents to leverage parameterized DQN to learn a comprehensive negotiation strategy that integrates linguistic communication skills and bidding strategies.

\subsubsection{Supervised Learning}
 
Supervised learning (SL) is another popular paradigm  for policy learning. \cite{lewis2017deal} adopt a Seq2Seq model to learn what action should be taken by maximizing the likelihood of the training data. However, supervised learning only aims to mimic the average human behavior, so~\citet{he2018decoupling} propose to finetune the supervised model to directly optimize for a particular dialogue reward function, which is defined as i) the utility function of the final price for the buyer and seller ii) the difference between two agents’ utilities iii) the number of utterances in the dialogue. \citet{zhou2019augmenting} train a strategy predictor to predict whether a certain negotiation strategy occurred in the next utterance using supervised training. The system response would be generated conditioned on the user utterance, dialogue context, and the predicted negotiation strategy. In addition, \citet{joshi2021dialograph} incorporate a pragmatic strategies graph network with the seq2seq model to create an interpretable policy learning paradigm.  Recently,  \citet{dutt2021resper} propose a generalised
framework for identifying resisting strategies in persuasive negotiations using a pre-trained BERT model~\cite{devlin2019bert}.




\subsubsection{Multi-task Learning}

Multi-task learning aims to jointly train the system with several sub-tasks and finally achieve satisfactory performance. \citet{li2020end} propose an end-to-end framework that integrates several sub-tasks including intent and semantic slot classification,  response generation and filtering tasks in a Transformer-based pre-trained model. \citet{zhou2019augmenting} propose to jointly model both semantic and strategy history using finite state transducers (FSTs) with hierarchical neural models. \citet{chawla2022opponent} integrate a preference-guided response generation model with a ranker module to identify opponents' priority.

%% file: 7-new_frontiers.tex
\section{New Frontiers and Challenges}
Previous sections summarize the prominent achievements of previous work in negotiation dialogue, including benchmarks, evaluation metrics and methodology. In this section, we will discuss some new frontiers which allow negotiation dialogue systems to be fit actual application needs and applied in real-world scenarios.

\subsection{Multi-modal Negotiation Dialogue}
Existing research works in negotiation dialogue only consider text format as inputs and outputs. However, humans tend to perceive the world in multi-modal patterns, not only text but also audio and visual information. For example, the facial expression and emotions of participants in a negotiation dialogue could be important cues for making negotiation decisions. Further work can consider adding this non-verbal information into the negotiation process.

\subsection{Multi-Party Negotiation Dialogue}
Although some work sheds light on multi-party negotiation, most current negotiation dialogue benchmarks and methods predominantly focus on two-party settings. Therefore, multi-party negotiation dialogues are under-explored. Future work can consider collecting dialogues in multi-party negotiation scenarios, including \emph{General multi-party negotiation} and \emph{Team negotiation.} Specifically, \emph{General multi-party negotiation} is a type of bargaining where more than two parties negotiate toward an agreement. For example, next-year budget discussion with multiple department leaders in a large company. \emph{Team negotiation} is a team of people with different relationships and roles. It is normally associated with large business deals and highlights the significance of relationships between multi-parties. There could be several roles, including leader, recorder, and examiner, in a negotiation team~\cite{halevy2008team}.

\subsection{Cross-Culture \& Multi-lingual Negotiation Dialogue}
Existing negotiation dialogue benchmarks overwhelmingly focused on English while leaving other languages and backgrounds under exploration. With the acceleration of globalization, a dialogue involving individuals from different culture backgrounds participants becomes increasingly important and necessary. That is, there is an urgent need to provide people with a negotiation dialogue system with different cultures and multi-lingual. Further works can consider incorporating multi-lingual utterances and social norms among different countries into negotiation dialogue benchmarks.


\subsection{Negotiation Dialogue in Real-world Scenarios}
As discussed in Section~\ref{datasetsection}, previous works have already proposed many negotiation dialogue benchmarks in various scenarios. However, we notice that most of these benchmarks are artefacts through human crowd-sourcing. Participants are often invited to play specific roles in the negotiation dialogue. The resulting dialogues may not perfectly reflect the negotiations in real-world scenarios (e.g., politics, business). Therefore, it could be a promising research direction to collect real negotiation dialogues. For example, one could collect recorded business meetings or phone calls.



%% file: 8-conclusion.tex
\section{Conclusion}

This paper presents the first systematic review on the progress of the negotiation dialogue system. We thoroughly summarize the existing works, which cover various domains and highlight their challenges respectively. Besides, we summarize currently available benchmarks, evaluations, and methodologies. In addition, we shed light on some new trends in this research field. We hope this survey can facilitate future research on negotiation dialogue systems.